# A Novel Self-Supervised Learning-Based Anomaly Node Detection Method Based on an Autoencoder in Wireless Sensor Networks

Miao Ye, Qinghao Zhang, Xingsi Xue, Yong Wang, Qiuxiang Jiang, Hongbing Qiu

*Abstract*—Due to the issue that existing wireless sensor network (WSN)-based anomaly detection methods only consider and analyze temporal features, in this paper, a self-supervised learning-based anomaly node detection method based on an autoencoder is designed. This method integrates temporal WSN data flow feature extraction, spatial position feature extraction and intermodal WSN correlation feature extraction into the design of the autoencoder to make full use of the spatial and temporal information of the WSN for anomaly detection. First, a fully connected network is used to extract the temporal features of nodes by considering a single mode from a local spatial perspective. Second, a graph neural network (GNN) is used to introduce the WSN topology from a global spatial perspective for anomaly detection and extract the spatial and temporal features of the data flows of nodes and their neighbors by considering a single mode. Then, the adaptive fusion method involving weighted summation is used to extract the relevant features between different models. In addition, this paper introduces a gated recurrent unit (GRU) to solve the long-term dependence problem of the time dimension. Eventually, the reconstructed output of the decoder and the hidden layer representation of the autoencoder are fed into a fully connected network to calculate the anomaly probability of the current system. Since the spatial feature extraction operation is advanced, the designed method can be applied to the task of large-scale network anomaly detection by adding a clustering operation. Experiments show that the designed method outperforms the baselines, and the F1 score reaches 90.6%, which is 5.2% higher than those of the existing anomaly detection methods based on unsupervised reconstruction and prediction. Code and model are available at https://github.com/GuetYe/anomaly_detection/GLSL

## I. INTRODUCTION

WIRELESS sensor networks (WSNs) are collections of densely deployed sensor nodes with event detection, data detection and communication capabilities. Each sensor node of a WSN mounts a large number of sensors for pressure, temperature and other physical quantities. Every sensor node can detect or respond to any event in the specified area and then send its values to nearby devices through a multihop wireless communication path for further processing [1]. WSN-based applications are used in many fields, such as smart homes, national defense, natural environment monitoring [2], health care services, and industrial production monitoring [3-4].

WSNs are mostly deployed in unattended areas, so the nodes are often vulnerable to attack. These attacks affect the normal operations of WSNs and cause problems such as wireless media interference, deployment environment deterioration, and transmission message interception. WSNs also have energy, memory, bandwidth and communication limitations, and when these limitations are broken, anomalies are caused. When an anomaly occurs in a WSN, the captured data flow deviates from the normal data distribution; exampled include contextual anomalies, point anomalies, and collective anomalies [5-6]. Specifically, point anomalies are data points that are unusual compared to the rest of the data, contextual anomalies are data points that are unusual compared to the normal data associated with them, and collective anomalies are collections of data points in several consecutive time intervals that are abnormal relative to the entire dataset. Considering that an individual data point may appear to be normal, the occurrence of several of these individuals at the same time constitutes a collective anomaly. In summary, anomalies can occur at any time in a WSN, and it is necessary to detect the abnormal states of WSNs in time and utilize emergency measures to prevent deterioration. This indicates that anomaly detection plays an essential role in the stable and safe operations of WSN systems.

Most existing abnormal node detection studies concerning sensor networks do not comprehensively consider the impact of the spatial location information between multiple sensor nodes and the correlation information between multiple modes for anomaly detection. For example, reference [7] only discussed the multimodal data flow of a single node, and its input was a matrix composed of multimodal time series measurement data. This work did not use the spatial location features of sensor nodes for anomaly detection. Reference [8] only discussed the multiple-node data flow of a single mode, and its input was a matrix composed of the time series data of multiple sensor node with a certain measurement mode, which did not take the correlations between different modes into account.

Miao Ye, Qinghao Zhang, Yong Wang, Qiuxiang Jiang, Hongbing Qiu are with School of Information and Communication, Guilin University of Electronic Technology, Guilin 541004, China(e-mail: yemiao@guet.edu.cn; boitha@foxmail.com; ywang@guet.edu.cn; jiangqiuxiang@guet.edu.cn; qiuhb@guet.edu.cn).

Xingsi Xue is with Fujian Provincial Key Laboratory of Big Data Mining and Applications, Fujian University of Technology, Fuzhou, Fujian, 350118, China (e-mail: jack8375@gmail.com).

In addition, due to the lack of labels in WSNs datasets, most of the existing works design anomaly detection methods via autoencoder reconstruction or use self-supervised learning for adversarial training, but these approaches do not comprehensively consider the advantages of traditional reconstruction-based and self-supervised learning-based anomaly detection models. In this paper, we propose a novel self-supervised learning-based anomaly node detection framework based on an autoencoder in a WSN.

The main contributions of this paper can be summarized as follows.

1) Sensor nodes at different locations in a WSN collect data from multiple different modalities at the same time to generate multimodal data flows at different spatial locations, and these data flows have strong correlations in space and time. Compared with the existing anomaly detection methods that only consider the multimodal time series data of a single node or the multinode time series data of a single mode, which fail to comprehensively consider the spatial and temporal features of multiple sensor nodes, the anomaly detection framework designed in this paper considers the temporal features of multimodal multinode data flows and the spatial features between multinode locations.

2) Compared with the existing anomaly detection framework based on autoencoder reconstruction, this paper introduces a temporal feature extraction module for the nodes in the local space, a Graph Neural Network (GNN)-based spatial and temporal feature extraction module for the nodes in the global space, and an adaptive fusion module in the encoding and decoding processes to learn the common and special features of nodes [9]. In the design of the optimization strategy for the training process, this paper combines the efficient and regular data feature capture capacity of the reconstruction-based model and the graph representation generalization ability improvement brought by self-supervised learning and proposes a novel anomaly detection framework based on the adaptive fusion of node spatial features and the data flow temporal features of the nodes in the **G**lobal and **L**ocal spaces via **S**elf-supervised **L**earning (GLSL).

3) The framework proposed in this article is a GNN "wrapper", which supports any kind of GNN kernel for completing the WSN anomaly detection task. In the case when a large-scale WSN faces a large number of sensor nodes, this paper divides all nodes into multiple clusters, the data flows of different clusters are trained and inferred on different anchor node devices, and dimensionality reduction is also carried out to reduce the model operation frequency, which can effectively reduce the hardware overhead and time consumption of the network.

The rest of this paper is organized as follows. Section II presents the background and related work on anomaly detection. Section III describes the basic knowledge of attribute networks and GNNs. Section IV describes the proposed self-supervised learning-based anomaly node detection framework built on an autoencoder in a WSN. We conduct extensive experiments to evaluate our method in Section V. Finally, Section VI concludes this paper.

## II. RELATED WORK

In recent years, researchers have proposed many methods for solving anomaly detection problems posed by WSNs. The authors in [10] used Euclidean distances to measure the similarity of WSN data points, divide the points into many clusters, and distinguish normal data from abnormal data according to the distances between the data points and cluster centers. Similarly, this method does not extract the correlations between the spatial locations of sensor nodes for anomaly detection. [11] proposed a multinode data flow anomaly detection method based on correlation analysis to extract the correlations of high-dimensional time series data for anomaly detection and to save computing costs. However, this method does not consider the multimodal scenarios of WSNs and does not use the correlation between multiple different modes for anomaly detection. In fact, each sensor node measures multiple modes in a WSN.

As the scale and perceived traffic of WSNs increase, the data flows of multiple data sources continue to be generated, and attack scenarios become more diverse and complex. While the existing WSN anomaly detection methods have achieved many good results, they have the problems of poor generalization and poor capacity to process high-dimensional, diverse, large-scale, and unbalanced data. To solve the above problems, deep learning technology [12] has been applied to anomaly detection systems. [13] studied an unsupervised multivariate anomaly detection method based on a generative adversarial network (GAN), which uses an LSTM network as the basic model for the generator and discriminator in the GAN framework to capture the temporal correlations of different data flows. [14] designed an anomaly detection method based on a variational autoencoder (VAE) and LSTM. [15] proposed an anomaly detection framework based on a CNN and incremental learning, solving the problem that complete retraining is required when new data come or anomaly classes are added. [9] studied an adaptive memory fusion network for anomaly detection via self-supervised learning, which adaptively fuses global and local memory features to learn common features and special features. [16] proposed an urban flow prediction method based on semantic representation learning, which captures the spatial features of different scales through semantic flow coding and employs a multihead self-attention mechanism to predict city flows. [17] used a minimax strategy to magnify the difference between abnormal data and normal data and utilized an attention mechanism to capture the difference between prior association and series association for anomaly detection. However, these methods do not use the spatial information of the network topology for anomaly detection, which limits the anomaly detection performance of the resulting models.

How the spatial features of the topology information possessed among the nodes of a WSN are applied to anomaly detection is key to achieving improved detection performance. GNNs can extract "interesting" features from the adjacency matrices and attribute matrices of attribute networks and can be utilized for any application based on a graph structure, so they have the ability to reflect the spatial characteristics of WSN data. Common GNNs include graph convolution networks (GCNs) [18], graph attention networks (GATs) [19], and the graph sampling and aggregation network (GraphSage) [20]. GNNs are widely used in graph representation learning. The goal of graph representation learning is to extract the "interesting" features of a graph to find a low-dimensional representation of the high-dimensional graph structure. [21] studied a graph embedding method called (GNN with LAbeling trickS for Subgraph (GLASS) based on a label strategy, which learns the features of the inner and outer subgraphs by assigning different labels and achieves better performance in the downstream tasks of GNNs. The convolution kernels in CNNs inspired the GNN-AK graph embedding framework of [22], which divides an entire large graph into multiple subgraphs and uses GNN kernels to extract the features of each subgraph, and the star topology of the traditional GNN aggregation method was also changed to egonet, which can better distinguish nonisomorphic graphs. Graph embedding technology has been widely used in various downstream tasks, including anomaly detection with graph structures. The cost of obtaining a large amount of labeled anomalous data for deep learning model training is very expensive, and normal state data are relatively easy to obtain. For this case, unsupervised learning, which only requires normal samples for training an anomaly detection model, is widely used to design anomaly detection methods. Unsupervised anomaly detection methods can be mainly divided into two classes according to their optimization strategies: reconstruction and prediction approaches. [23] used popular GCNs for node embedding learning. Then, the learned embeddings were used to reconstruct the original data, and the current system state was judged as abnormal when the reconstruction error exceeded the threshold. The authors in [24] changed their GCN to a GAT on the basis of [23], thereby solving the problem of oversmoothing. The authors in [25] combined the structure learning method with a GNN and used historical data to predict the subsequent sensor observations, and the system state was regarded as abnormal when the prediction error exceeded the threshold. [7] combined the advantages of forecast-based models and reconstruction-based models and proposed a novel anomaly detection framework to obtain better data representations by combining the prediction of the next timestamp data and the reconstruction of the entire data flow. [26] studied an anomaly detection method based on a residual GCN, which uses a residual attention mechanism to reduce the adverse effects induced by abnormal nodes and prevent excessive smoothing and calculates an anomaly score through a residual matrix to judge the system state. Since the traditional anomaly detection methods based on unsupervised reconstruction that are trained on limited normal samples tend to overfit the data, to solve this problem, many researchers have proposed the use of self-supervised learning to improve the generalization abilities of neural network models. The author in [27] designed a two-stage adversarial training-based anomaly detection model. The first-stage model is committed to reconstructing the input data, and the second-stage model is used as a generator for generating a negative sample to fool the discriminator. Furthermore, the discriminator needs to distinguish the generated negative samples from the positive samples. The authors in [28] studied an anomaly detection method combining the deep graph infomax (DGI) [29] and GraphSage techniques; their approach samples subgraphs, constructs positive and negative samples, and then uses a discriminator to conduct adversarial training with the positive and negative samples. The authors in [30] studied an anomaly detection framework called Dynamic-DGI that uses the mutual information between corrosion subgraphs and the total input graph for adversarial training. [31] proposed an anomaly detection framework based on multiscale adversarial training. It uses a GNN to construct two adversarial learning networks for learning the distribution of the raw data possessed by nodes at different scales, and uses the consistency of the test data at multiple scales to evaluate the states of nodes. However, the data flow generated by a WSN can be regarded as multiple sensor nodes with a certain topology structure that collects the dynamic time series data of multiple modes with respect to the surrounding environment. The data flows include not only the topology information of spatial locations but also the time series temporal information of multiple modes. In other words, WSN data in the real world are data flows containing three dimensions (nodes, modes, and time), and the anomaly detection methods described above do not consider the features of these three dimensions simultaneously. They consider only two dimensions, such as a matrix composed of multimodal readings obtained over a period of time or a matrix composed of the multinode values of a certain measurement attribute (i.e., a single mode) over a period of time. These approaches do not comprehensively consider the correlations between times, modes and spatial node positions when conducting anomaly detection for WSN data flows.

To make full use of the correlation information located between node positions and different modes in a WSN for anomaly detection, [32] first processed the multimodal data flows of each node separately, extracted the correlation features between different modes and different moments, then extracted the spatial features of the WSN according to the spatial positions of the nodes to predict the WSN data at the current moment, and finally used the prediction error and a threshold to determine the WSN state. However, the drawback of this method is that each node generates a branch, and the number of branches expands rapidly when the number of sensor nodes increases, so the calculation speed is problematic for large-scale WSN scenarios. Moreover, the method is a prediction-based model, which cannot capture the distribution features of entire time series data as reconstruction-based models can, and it becomes increasingly difficult to make accurate predictions over time, which affects the anomaly detection performance of the method. To solve this problem, we carry out follow-up work based on a GNN. This paper combines the efficient and regular data feature capture capacity of a reconstruction-based model and the advantage of improving graph representation generalization ability improvement brought by self-supervised learning, fusing the temporal features of the global location nodes and local location nodes of a WSN to comprehensively determine its system state. This improves the detection performance and reduces the time consumption and space overhead of the resulting model.

### III. FUNDAMENTAL METHODS

*A. Problem definition*

Before discussing the problem definition for the anomaly detection framework, we first explain the spatial position correlations between the sensor nodes and the temporal correlations between data flows with different modes [33]. The spatial position correlations between the sensor nodes mean that the data collected by any sensor node and its neighboring nodes within a short spatial distance also have a certain correlation. As shown in Fig. 1, each solid circle represents a sensor node. If there is a fire near sensor node $o$ and $o$ is sufficiently close to $a$, $b$ and $c$, the fire also affects the readings of nodes $a$, $b$ and $c$ (e.g., the temperature values substantially increase).

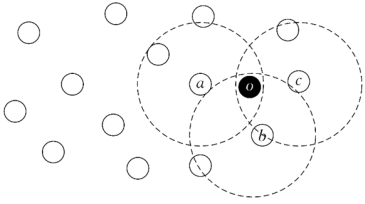

**Fig. 1.** Different nodes monitor the same event source. Each solid circle represents a sensor node. If a fire occurs near sensor node $o$ and $o$ is sufficiently close to $a$, $b$, and $c$, the fire also affects the readings of nodes $a$, $b$, and $c$ to some extent. This indicates that different data flows from different sensor nodes at different locations have spatial correlation features.

The temporal correlations between data flows with different modes means that the multimodal data flows collected by a sensor node are not isolated, and the data flows collected by any mode and other modes also have certain correlations. As shown in Fig. 2, when a fire occurs, the temperature and carbon dioxide concentration readings increase, while that of humidity falls, which means that the temperature value is positively correlated with the $CO_2$ concentration and negatively correlated with humidity.

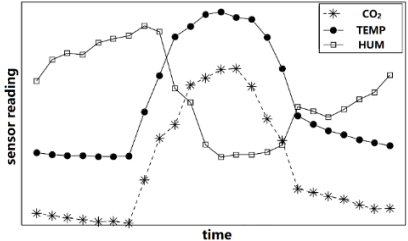

**Fig. 2.** Correlation features between multiple modes. The figure shows the time-varying curves of three modal values measured on a sensor node in a fire event. When a fire event occurs, the temperature and carbon dioxide values rise, while the humidity value falls. This illustrates that different data flows in one sensor node have time correlation features.

This paper aims to make full use of the spatial position correlations and the temporal correlation features described above for anomaly detection and models the anomaly detection problem of WSN data flows as a classification problem. Assume that there are $N$ nodes in a WSN, and each node measures $M$ modes. Then, the observation of the sensor network at time $t$ can be recorded as $x_t \in R^{M \times N}$. Assuming that the historical window size considered for each operation is $W$, the sensor network observations within $W$ historical consecutive moments form the input $X_t = \{x_{t-W+1}, x_{t-W+2}, ..., x_t\} \in R^{M \times N \times W}$ for the current moment $t$. The output $y = f(X_t | \theta) \in \{0,1\}$ of the anomaly detection task is the state label indicating whether the current moment is an anomaly or not, where $f$ is a mapping function, $\theta$ denotes the parameters to be estimated by the model $f$, and $y = 1$ indicates that the WSN state is anomalous ($y = 0$ if not). In real WSN scenarios, the lack of ground-truth labels often hinders the classification task, and the amount of abnormal data in WSNs is far less than the amount of normal data. Therefore, we use the self-supervised learning method to artificially inject anomalies into WSN data to generate negative samples and then use a GNN to build a deep learning model for classifying positive and negative samples.

*B. Attribute network*

Suppose that we are given an attribute network $G=(A, X)$, in which $A \in R^{M \times M}$ represents the adjacency matrix of the network, $M$ represents the total number of nodes in the network and $X \in R^{M \times D}$ represents the attribute matrix of $M$ nodes. The attribute features of each node can be represented by a $D$-dimensional vector, and the $D$-dimensional vectors of the $M$ nodes are combined to form the attribute matrix $X$ of the attribute network.

*C. GNN*

The proposed GLSL framework acts as a "wrapper" for any base GNN kernel to achieve the WSN anomaly detection task, so we need to introduce the GNN first.

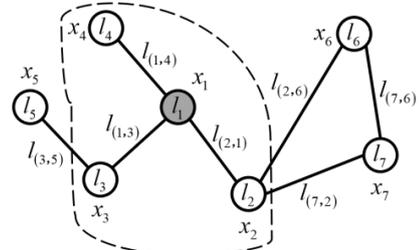

**Fig. 3.** Graph $G$ and central node 1. Node 2, node 3 and node 4 are the neighboring nodes of node 1, $l_i$ represents the representation vector of node $i$, $l_{(i,j)}$ represents the representation vector of the edge between nodes $i$ and $j$, and $x_i$ represents the current state of node $i$.

Reference [34] studied a GNN as early as 2009. The essence of a GNN is to use the central node features, the features of the neighbor nodes of the central node, and the features of the edges between the central node and the neighbor nodes to generate the current state of the central node. In other words, a GNN uses graph structural topology information to aggregate the neighbor node features of the central node to generate the state of the central node. Fig. 3 shows a graph $G = (N,E)$ and its central node $l$, where $N$ is the node set and $E$ is the edge set. $l_i$ is the label of node $i$, $l_{(i,j)}$ is the label of the edge between nodes $i$ and $j$, and $x_i$ represents the current state of node $i$. Then, $x_i$ and the output $O_i$ are defined as follows:

$$x_i = f_w\left(l_i, l_{co[i]}, x_{ne[i]}, l_{ne[i]}\right) \qquad (1)$$

$$O_i = g_w\left(x_i, l_i\right) \qquad (2)$$

where $l_{co[i]}$ is the label collection of all the edges connected to node $i$, $x_{ne[i]}$ is the state collection of all neighboring nodes of node $i$, and $l_{ne[i]}$ is the label collection of all neighboring nodes of node $i$. $f_w$ is a propagation function that reflects the dependency of the current node $i$ on its neighboring nodes. $g_w$ is an output function that can convert the states and labels described above into the final output of node $i$. The description of (1) is difficult to understand and abstract, and we can convert (1) into (3) according to Fig. 3.

Node $l$ combines its own labels, all labels of the edges connected to it, and the states and labels of the neighboring nodes to generate its current state.

$$x_1 = f_w\left(l_1, \underbrace{l_{(2,1)}, l_{(1,3)}, l_{(1,4)}}_{l_{co[1]}}, \underbrace{x_2, x_3, x_4}_{x_{ne[1]}}, \underbrace{l_2, l_3, l_4}_{l_{ne[1]}}\right) \qquad (3)$$

Equations (1-2) can be represented as (4-5) by the classic iteration method, where $x_i(t)$ represents the state of the $i^{th}$ node at the $t^{th}$ iteration, (4) shows the state transition from the $t^{th}$ iteration to the $t+1^{th}$ iteration, and (5) describes the transformation of the state and label of the $i^{th}$ node at the $t^{th}$ iteration into the final output.

$$x_i(t) = f_w\left(l_i, l_{co[i]}, x_{ne[i]}(t-1), l_{ne[i]}\right) \qquad (4)$$

$$O_i(t) = g_w\left(x_i(t), l_i\right) \quad i \in N \qquad (5)$$

C. GAT

A GAT is a well-known GNN variant. As shown in Fig. 3, a GAT calculates the attention weight $a_{ij}$ between the central node $h_i^{(l)}$ and each neighboring node $h_j^{(l)}$ according to their correlation based on an attention mechanism and adaptively assigns weights to different neighbors, greatly improving upon the expression ability of the GNN model and solving the oversmoothing problem of multilayer GCNs.

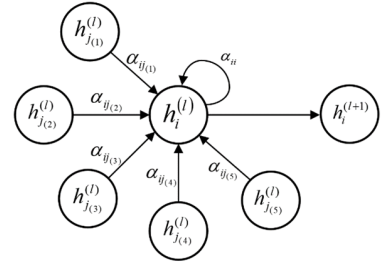

**Fig. 4.** GAT calculation process. The central node $h_i^{(l)}$ aggregates the information of the neighboring nodes [ $h_{j(1)}^{(l)}, ..., h_{j(5)}^{(l)}$ ] through the allocated attention weight coefficient at the $l^{th}$ layer, and its own information is also taken into consideration.

Assuming that the given GAT has $L$ layers, the representation of node $i$ in the $l^{th}$ layer is $h_i^{(l)} \in R^{F^{(l)}}$, where $F^{(l)}$ is the dimensionality of the representation of node $i$ in the $l^{th}$ layer. Assume that the total number of nodes is $N$, $H^{(l)} = \left\{h_1^{(l)}, h_2^{(l)}, ..., h_N^{(l)}\right\}$ is the input of the $l^{th}$ layer, and the output of the $l^{th}$ layer is $H^{(l+1)} = \left\{h_1^{(l+1)}, h_2^{(l+1)}, ..., h_N^{(l+1)}\right\}$. The representation of node $i$ in the $l+1^{th}$ layer is $h_i^{(l+1)} \in R^{F^{(l+1)}}$, where $F^{(l+1)}$ is the dimensionality of the representation of node $i$ in the $l+1^{th}$ layer. Note that the representation at the $0^{th}$ layer $H^{(0)} = X = \{x_1, x_2, ..., x_N\}$ is the attribute matrix of the attribute network.

First, we use a weight matrix $B \in R^{F^{(l+1)} \times F^{(l)}}$ to map the representation vector of the input node $i$ in the $l^{th}$ layer to the hidden layer vector $q_i^{(l)}$.

$$q_i^{(l)} = Bh_i^{(l)} \qquad (6)$$

Second, the correlation coefficient between node $i$ in the $l^{th}$ layer and node $j$ in the $l^{th}$ layer can be computed as in (7).

$$e_{ij}^{(l)} = Leaky\text{Re}LU\left(a^{(l)^T}\left[q_i^{(l)} \| q_j^{(l)}\right]\right) \qquad (7)$$

where $a^{(l)} \in R^{2F^{(l+1)}}$ is the weight vector in the $l^{th}$ layer and node $j$ is the neighbor node of node $i$. From the adjacency matrix $A$, we can easily obtain the full neighbor node information of a certain node. $\|$ represents the combination operation. $e_{ij}^{(l)}$ indicates the correlation coefficient between node $i$ and node $j$ in the $l^{th}$ layer. Then, the correlation coefficient calculated between each node and all related neighbors is normalized by a *softmax* function to obtain the attention weight matrix.

$$\alpha_{ij}^{(l)} = \exp\left(e_{ij}^{(l)}\right) / \sum_{v_k \in N(v_i)} \exp\left(e_{ik}^{(l)}\right) \qquad (8)$$

where $\alpha_{ij}^{(l)}$ represents the attention weight coefficient between node $i$ and node $j$ in the $l^{th}$ layer. $N(v_i)$ represents the set of neighboring nodes of node $i$. The normalization process in (8) ensures that the sum of the attention weight coefficients between every node and all its neighboring nodes is 1. Finally, the representation $h_i^{(l+1)} \in R^{F^{(l+1)}}$ of node $i$ in the $l+1^{th}$ layer is calculated using the attention weight coefficients and a hidden layer vector, where $\sigma$ is a nonlinear activation function.

$$h_i^{(l+1)} = \sigma\left(\sum_{j \in N_i} \alpha_{ij}^{(l)} q_j^{(l)}\right) \quad (9)$$

In practical applications, to make the effect of self-attention more stable, the representation of the $l+1^{th}$ layer can be generated by using multihead attention. Equations (6-9) describe the iterative formula of a single attention head from the $l^{th}$ layer to the $l+1^{th}$ layer, and the multihead attention generalizes one attention head to $K$ attention heads, as shown in (10), where $\|$ represents the stitching operation, $\alpha_{ij}^{(l)}(k)$ represents the attention coefficient between node $i$ and node $j$ of the $k^{th}$ attention head, and $q_j^{(l)}(k)$ represents the hidden layer representation of node $j$ in the $k^{th}$ attention head. Through k-head attention processing, the $F^{(l)}$-dimensional input of the original node $i$ is transformed into the $k*F^{(l+1)}$ dimension.

$$h_i^{(l+1)} = \mathop{\|}\limits_{k=1}^{K} \sigma\left(\sum_{j \in N_i} \alpha_{ij}^{(l)}(k) q_j^{(l)}(k)\right) \quad (10)$$

*D. Topology construction*

Assume that $N$ sensor nodes are randomly distributed in the rectangular space of size $R$x$R$, and each sensor node has a unique identifier and specific position coordinates ($P_x$, $P_y$), where $P_x$ and $P_y$ are the $X$-axis coordinate and $Y$-axis coordinate of the sensor node in two-dimensional space, respectively, $0 \le P_x \le R$, and $0 \le P_y \le R$. In this paper, every sensor node is considered a node of the graph, and the following three methods are introduced to construct the WSN topology.

1. Complete graph: The complete graph is used to construct the adjacency matrix of the WSN, which means that each sensor node regards all other sensor nodes as neighbor nodes and constructs edges with them. When the monitoring area covered by the WSN is smaller than the event impact range, the influence of spatial location on the observations is weak, and the nodes can communicate with each other, it is more reasonable to use this topology construction method.

2. Coverage distance: Assume that each sensor node can only cover a circular area with a radius $cd$ centered on itself; that is, the communication radius of each sensor node is $cd$. At this time, any other sensor node within the communication radius of the current sensor node is regarded as a neighbor node, and then an edge is constructed between the current node and the neighbor node.

3. TopK: The adjacency matrix of the WSN is constructed by the TopK method. Any sensor node can only construct edges with the TopK nodes closest to it in space based on their Euclidean distances and does not construct edges with other sensor nodes.

IV. PROPOSED ANOMALY DETECTION METHOD

*A. Multimodal WSN data flow model considering temporal and spatial features*

When discussing the correlation features between the different nodes and different modes of a data flow, a WSN data flow model should be established first. Assume that a WSN is deployed in a certain target area that contains $N$ sensor nodes and that $M$ sensors are deployed in each node to monitor the region's temperature, humidity, carbon dioxide concentration and other indicators. A time synchronization mechanism can be used to ensure the synchronization of the information transmission and data collection among the nodes in the WSN.

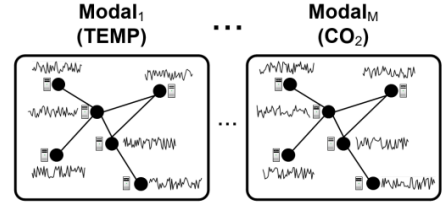

**Fig. 5.** Multinode multimodal WSN data flow. There are multiple nodes in the space that measure the time series data of multiple modes, and $M$ is the number of modes (e.g., temperature or humidity) measured by each sensor node.

To make full use of the historical data and current observation data for the real-time analysis of data features, we adopt a sliding window model to build a data flow model [35]. Assuming that the current timestamp is $t$ and that the length of the window is set to $W$, the observations of all sensors in all sensor nodes within the time step $\{t-W-1, t-W,..., t-1\}$ are truncated to form a data flow tensor $X \in R^{M \times N \times W}$. The data of the leftmost element expire when a new data element enters from the right, and the window slides by a length of one data element to the right. For example, the time range of the data flow model at time $t$ is $\{t-W-1, t-W,..., t-1\}$, and the time range of the data flow model at time $t+1$ is $\{t-W, t-W+1,..., t\}$. Therefore, the WSN data flow can be regarded as a dynamic network that changes with time. Fig. 5 shows a visual model of the WSN data flow, in which each node can collect time series data with respect to multiple attributes (modes), such as temperature (temp), carbon dioxide concentration ($CO_2$), and oxygen concentration ($O_2$).

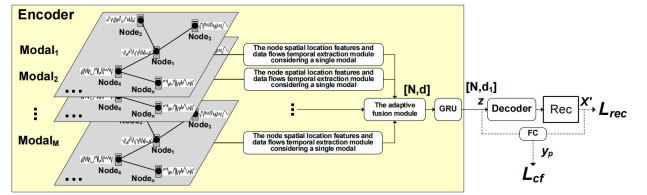

**Fig. 6.** GLSL framework. The input data are the multinode multimodal time series data, the input is converted into the hidden layer $Z$ through the encoder, and $Z$ is fed into the decoder to obtain the reconstruction input data $X'$. Eventually, the reconstruction result and the hidden layer representation of the encoder are fed into a fully connected network to calculate the anomaly probability of the current system.

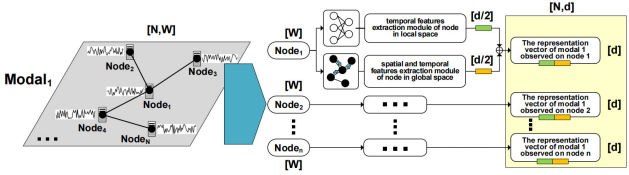

**Fig. 7.** The module for extracting node spatial location features and temporal data flow features considering a single mode. The input data are the WSN data flows of multiple nodes considering only a single mode. By extracting the temporal features of the nodes in the local space, the temporal features of the nodes in the global space and the spatial position features, the multinode feature representation matrix of a single mode is generated.

*B. The designed WSN anomaly node detection model*

Fig. 6 shows the designed WSN anomaly node detection model. We use a GNN to construct an autoencoder network as the basic network of the anomaly detection framework, and the autoencoder consists of an encoder $f_e$ and a decoder $f_d$. $\theta_e$ and $\theta_d$ are the relevant parameters of $f_e$ and $f_d$, respectively. As shown in (10-11), the encoder $f_e$ converts the input WSN data flows into a hidden layer representing $Z$, and $d$ is the hidden layer dimensionality parameter, which is set by the user. Then, the decoder can reconstruct an input $X'$ with the same shape by mapping $Z$. Finally, the hidden layer representing $Z$ and the reconstruction result $X'$ are fed into the fully connected network (i.e., *FC*) and the *softmax* function to jointly calculate the classification probability (i.e., the probability of belonging to the normal class or the abnormal class). The formula for this process is as follows:

$$Z = f_e(X;\theta_e),\ \ X' = f_d(Z;\theta_d) \quad (11)$$

$$y = \text{Soft}\max(FC(Z,X')) \quad (12)$$

In this paper, the architecture of the designed framework is trained in two phases. In the first stage, the model needs to capture the data distribution of the entire time series and reconstruct the input normal WSN data flows. In the second stage, the input includes the WSN data flows with injection anomalies, and the model should accurately classify the WSN data at normal and abnormal moments. The specific details are introduced in the following sections.

*C. Preprocessing module*

Different measurement attributes (modals) often have different measurement ranges, so the orders of magnitude between different modes vary greatly. If the original data are directly used for analysis purposes, the roles of modes with higher values in the comprehensive analysis are strengthened, and the roles of modes with lower values in the comprehensive analysis are weakened. Therefore, to eliminate the influence of the measurement range differences between modes, the original data must be standardized such that each mode has the same value range. In this paper, $Z$ score standardization is used to standardize the input tensor. The equation for doing so is as follows:

$$\widetilde{X}_{ij} = \left(X_{ij} - \mu(X_{ij})\right)/\sigma(X_{ij}) \quad (13)$$

where $X \in R^{M \times N \times W}$ is the input tensor, $X_{ij}$ represents the data flow measured by the $i^{th}$ sensor of the $j^{th}$ node, $\mu(X_{ij})$ represents the mean value of $X_{ij}$ and $\sigma(X_{ij})$ represents the standard deviation of $X_{ij}$. After $Z$ score preprocessing, the multimodal data flows conform to a normal distribution with a mean of 0 and a standard deviation of 1.

*D. The encoder and decoder based on node spatial location feature extraction, temporal feature extraction and multimodal correlation feature extraction*

The encoder is mainly composed of two modules: The module for extracting node spatial location features and temporal data flow features considering a single mode, and an adaptive fusion module.

The modes in the WSN data are layered, each mode constructs a branch, and the input data of each branch form a matrix $X_i \in R^{N \times W}$ composed of the multinode data flows of the current considered mode $i$. First, the input data are processed by the model for extracting the node spatial location features and temporal data flow features considering a single mode to extract the spatiotemporal features of the sensor nodes with respect to the current mode and generate the representations of all nodes with respect to the current mode. After all branches are processed, the node representations obtained from all modal branches are adaptively fused to extract the correlation features between the modes. Second, a gated recurrent unit (GRU) is used to solve the long-term dependency problem and obtain the hidden layer representation of the encoder. Finally, the original input is reconstructed through the decoder. The structure of the decoder is highly similar to that of the encoder, and it can be regarded as a mirror image of the encoder, so we omit its description. The two modules described above are introduced in detail as follows.

1) *The module for extracting node spatial location features and temporal data flow features considering a single mode*

Fig. 7 shows the process of each branch in detail. The module for extracting node spatial location features and temporal data flow features considering a single mode is mainly composed of a temporal feature extraction module for the nodes in local space and a spatial and temporal feature extraction module for the nodes in the global space.

According to the discussion concerning the extraction of features from time series data in [9], fully connected neural networks can be used to extract the temporal features of time series data flows, so the temporal feature extraction module designed in this paper for the nodes in local space is composed of a fully connected network. The WSN data $X_{ij}$ of mode $i$ observed on node $j$ is a vector with $W$ dimensions, and the temporal features of $X_{ij}$ can be extracted and reduced to a $d/2$-dimensional vector through the fully connected layer. Assuming that the total number of layers of the fully connected network is $K$, Equations (14-16) show the iterative process of the fully connected network. where $H^k$ is the hidden layer representation of the fully connected network at the $k^{th}$ layer, $H^0$ is the data vector of mode $i$ on node $j$, $a^k$ and $b^k$ are the weights and biases of the fully connected network at the $k^{th}$ layer, respectively, and $g$ is the output function. In general, the temporal features of the input data flow $X_{ij} \in R^W$ are extracted through the fully connected network and converted into $X_{ij}^l \in R^{d/2}$ with $d/2$ dimensions. In the temporal feature extraction module for the nodes in local space, instead of acquiring the features of other nodes to generate a new representation, every node independently extracts its own unique special features through a fully connected network.

$$H^0 = X_{ij} \qquad (14)$$

$$H^k = \sigma\left(a^k H^{k-1} + b^k\right) \quad k \in 1, 2, ..., K \qquad (15)$$

$$X_{ij}^l = g\left(H^K\right) \qquad (16)$$

The spatial and temporal feature extraction module for the nodes in the global space is composed of a GNN, and the $i^{th}$ modal data flow observed on node $j$ is combined with the spatial topology of the WSN to aggregate the information derived from the neighboring nodes, extract more common global features from a global spatial perspective and generate a representation of node $j$ considering mode $i$, reducing the dimensionality to $d/2$. (17) defines the feature extraction process based on a GNN, where $X_{ij}$ is the data flow of the $i^{th}$ mode observed on node $j$, $u$ is the neighbor node of node $j$, $H_{ju}$ is the weight of the edge between node $u$ and node $j$, $\{X_{iu}\}$ is the data flow collection of all neighbor nodes of $j$, $\{H_{ju}\}$ is the weight collection of all $H_{ju}$ values, and $\Phi_{GNN}$ and $f_{GNN}$ are the update function and aggregation function of the GNN, respectively. In general, $X_{ij} \in R^W$ combines the information of the current node's neighbors according to the topology from a global spatial perspective to extract its common features and is converted into $X_{ij}^g \in R^{d/2}$ through a GNN. Since the designed framework can be said to be a GNN-based "wrapper", any GNN variant can be used to detect anomalies for WSN data flows through the framework proposed in this paper.

$$X_{ij}^g = \Phi_{GNN}\left(X_{ij}, f_{GNN}\left(\{X_{iu}\}, \{H_{ju}\}\right)\right) \quad u \in N(j) \qquad (17)$$

*2) The adaptive fusion module*

The adaptive fusion module aims to adaptively fuse the special and common features of the nodes extracted from different modes. First, the extracted special features and common features are spliced together, and then the spliced features between different modes are fused by a weight $q$ to obtain the representation $X_{:j}^c$ of node $j$. This process is shown in (18), where $q_i \in R^d$ is the fusion weight of the $i^{th}$ mode. These weights are randomly initialized and then trained along with the mode.

$$X_{:j}^c = \sum_{i=0}^{M} q_i \left(X_{ij}^g \| X_{ij}^l\right) \qquad (18)$$

To further extract the temporal features of WSN data and solve the long-term dependency problem, this paper further processes the WSN data through two GRU layers [36] to obtain the hidden layer representation of the encoder, where $h$ represents the previous hidden state. As described in Section IV-B, this paper uses an autoencoder network structure to reconstruct and classify the input data. Because the decoder structure is highly similar to the encoder structure and can be regarded as a mirror of the encoder, it is not specifically repeated. The hidden layer representation of the encoder reconstructs the original input $X'$ through the decoder, which is shown in (19-20).

$$Z = GRU\left(X^c, h\right) \qquad (19)$$

$$X' = decoder(Z) \qquad (20)$$

*E. Two-phase training based on data reconstruction and self-supervised learning*

The existing unlabeled dataset anomaly detection methods based on deep learning can be divided into two categories: unsupervised and self-supervised approaches.

Unsupervised methods: Unsupervised learning methods often reconstruct the input normal data or predict the values of the next moment, and the reconstruction error or prediction error is used to detect anomalies. An autoencoder is an artificial neural network that is commonly used in unsupervised learning tasks to learn efficient representations of input data. It encodes the original input through the encoder to generate a hidden layer representation and then reconstructs the original input through the decoder. In the above process, the autoencoder learns the distribution of the full input data by continuously optimizing its parameters to reduce the reconstruction error, and it can convert the original input into a reasonable and efficient representation (i.e., a hidden layer representation). When abnormal data that differ greatly from the normal data distribution are fed into the model, the reconstruction error surges.

Self-supervised methods: Anomaly detection models tend to overfit when conducting unsupervised training on a limited number of normal samples, and to solve this problem, many anomaly detection methods use self-supervised learning to train models. Self-supervised learning enables a model to learn the difference between normal and abnormal data by performing adversarial learning with generated negative samples and positive samples. The negative samples (i.e., pseudodata) can be generated by a GAN or other methods. In other words, self-supervised learning transforms an unsupervised problem into a supervised problem.

In this paper, the model of the designed framework is trained in two phases, combining the efficient normal data feature capture capacity of the reconstruction-based model and the advantage of improving the graph representation generalization ability via self-supervised learning. In the first stage, we train the model to learn the distribution features of normal input data. In the second stage, we construct negative samples by artificially injecting multiple anomalies, and the model needs to learn the ability to correctly distinguish the negative samples from the normal samples.

Stage 1: Reconstruction. In the first stage, the goal of GLSL is to train the model to reconstruct the input normal data, as shown in (21). The input WSN data $X$ are converted into a hidden layer representation $Z$ by the encoder, and then the input data are reconstructed by the decoder. Minimizing the reconstruction error is the optimization goal in this stage. The loss function is shown below, where $x_i$ and $x_i'$ are the sums of the elements in $X$ and $X'$, respectively, and $l$ is the total number of elements in $X$.

$$L_{rec} = \frac{1}{l}\sum_{i=0}^{l}(x_i - x_i')^2 \qquad (21)$$

Stage 2: Self-supervised learning. In this stage, the goal of GLSL is to train the model to learn the difference between normal data and abnormal data so that the model has the capacity to determine whether new data are abnormal. Assuming that a raw input $X$ is given, injecting an anomaly into a determined part of the continuous moments of the normal samples can generate negative samples. In the second stage, the goal of GLSL is to distinguish between abnormal moments with abnormal input data and normal moments with normal input data. In other words, when the abnormal injection moment enters the sliding window, GLSL needs to determine that the system is under an abnormal state; otherwise, it is judged to be normal. Minimizing the classification error is the optimization goal in this stage, and the loss function is shown in (22), where $y$ is the label and $y_p$ represents the prediction probability of a positive sample.

$$L_{ce} = -y\log(y_p) - (1-y)\log(1-y_p) \qquad (22)$$

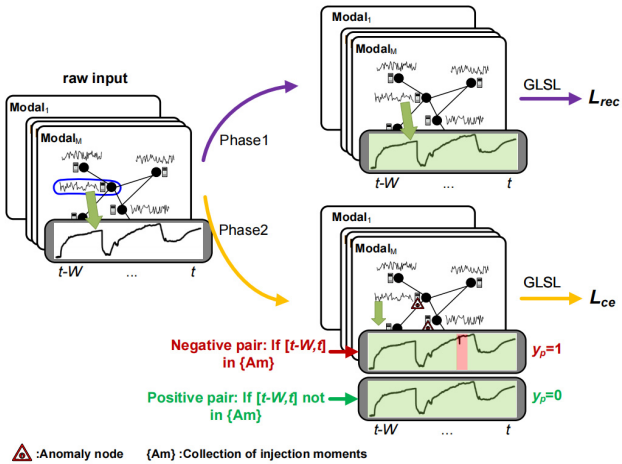

**Fig. 8.** Two-stage training. In the first stage, the task of the model is to reconstruct the normal input data. In the second stage, the task of the model is to correctly distinguish between the anomaly injection time and the normal time without anomaly injection. Due to the large numbers of nodes and modes, this figure only uses the data flow of a mode observed on a node for visual display purposes.

Fig. 8 shows the complete two-stage training procedure. Combining (21) and (22), the total GLSL training process can be defined as (23), where $n$ is the number of training epochs. For the inference process, the binary classification probability at the current moment is obtained by the classification output branch to determine the state of the current moment.

$$L = \frac{1}{n}L_{rec} + \left(1 - \frac{1}{n}\right)L_{ce} \qquad (23)$$

*F. Improving scalability: Strategy for large-scale WSN scenarios*

The sizes of WSN datasets in the real world are very large, and with the increase in the number of layers in GNNs, the computational cost and required memory space increase exponentially, so the computational and storage complexities encountered when training large-scale GNNs are very high. To cope with the high time consumption caused by the massive numbers of sensor nodes and recording moments in large-scale WSN scenarios, this paper proposes an expansion strategy called GLSL+ based on K-means and piecewise aggregate approximation (PAA).

K-means clustering is a well-known clustering algorithm that is concise and efficient, making it the most widely used clustering algorithm. Given a set of spatial node coordinates $C$ and the number of clusters $k$ ($k$ is a parameter set by the user), the k-means algorithm can repeatedly iterate according to the Euclidean distance measure and finally divide the adjacent nodes in terms of their spatial positions into different class clusters. PAA [37] is an efficient dimensionality reduction tool designed for large-scale data, and the original data can be reduced while retaining the features of the original data flow by using PAA. Suppose there is a primitive temporal data flow $Y = \{Y_1, Y_2, ..., Y_n\}$ of length $n$, which can be converted into a data flow $\overline{Y} = \{\overline{Y}_1, \overline{Y}_2, ..., \overline{Y}_m\}$ of length $m$ by (24).

$$\overline{Y}_i = \frac{m}{n}\sum_{j=n/m(i-1)+1}^{(n/m)i} Y_j \qquad m \leq n \qquad (24)$$

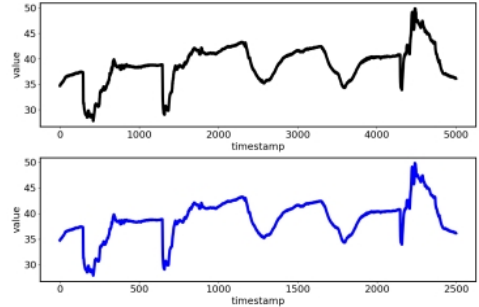

**Fig. 9.** PAA. The upper part of the figure is a data flow of 5000 timestamps with the humidity mode observed on node 5. After completing PAA processing, the 5000 timestamps are reduced to 2500 timestamps without losing sequence characteristics. The data obtained after dimensionality reduction are shown in the lower part of the figure.

As shown in Fig. 10, GLSL+ first divides the nodes of the entire dataset into $k$ clusters according to the spatial positions of the nodes through K-means. The distance from each cluster node to its corresponding cluster center is less than the distances to other cluster centers. Then, the time dimensionality is reduced through PAA, and the different data of different clusters are trained and inferenced on different devices. The splitting operation in GLSL+ for the node dimension can effectively reduce the parameters of the generated model and can be independently trained and inferred on different devices to achieve acceleration. The reduction in the time dimension can convert $D$ data points into fewer ($D_l$) input data points and reduce the information loss caused by the conversion process as much as possible to reduce the number of runs and achieve further acceleration.

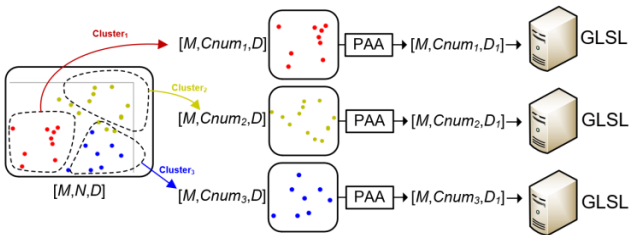

**Fig. 10.** GLSL$^+$ optimization strategy under a large-scale WSN.

## V. Experiments

### A. Experimental setup

To prove the feasibility of the anomaly detection method proposed in this paper, this section performs model training, anomaly detection and performance verification on WSN sensor data through experiments. The hardware environment used for the experiment is a server with an Intel Xeon Silver 4210 CPU @2.20 GHz and an NVIDIA GeForce RTX 2080Ti GPU. The software environment used for the experiment is Python. We implement our method and its variants in the PyTorch-1.9.0 deep learning framework and the Matplotlib drawing library. The version number of CUDA is 11.1.

### B. Dataset and anomaly injection method

In this paper, we use the WSN dataset of the Intel Berkeley Research Lab (IBRL) [38] to verify the effectiveness of our model. The IBRL sensor network is composed of 54 sensor nodes, each of which collects four modal attributes: temperature, humidity, illumination strength and voltage. The data collection period ranged from 28 February 2004–5 May 2004, and the spatial locations of the sensor nodes are shown in Fig. 11.

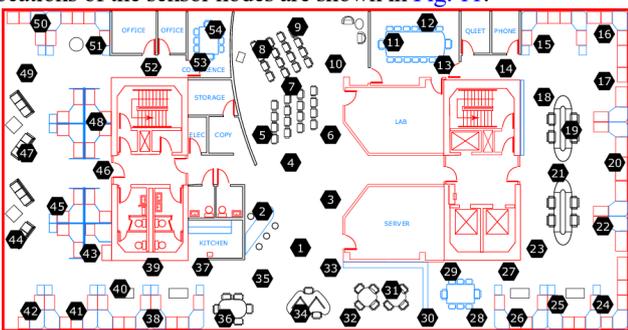

**Fig. 11.** Sensor node deployment in the IBRL dataset.

In the experiment, we select three modalities of data (temperature, humidity and voltage) from 51 sensor nodes at 5000 moments from March 4 to March 9 to form the original data tensor with a shape of 50 × 3 × 5000. The ratio of the training set to the test set is 6:4, so the training set contains 3000 sampling moment data with three modalities from the 51 sensor nodes, and the test set contains 2000 sampling moment data with three modalities from the 51 sensor nodes. As described in Section V-E, GLSL uses self-supervised learning by constructing negative samples through the injection of anomalies in the second stage, and the five anomaly detection methods are detailed as follows.

1. Scale change: The values of the WSN data in the sliding window are changed by multiplying them by a random constant. Here, we choose [0.5, 1.5, 2] as the random constants. The addition of scale transformation to self-supervised learning can help the model learn the features of scale changes in the WSN data so that the model becomes sensitive to abnormal scale changes in the data.

2. Negation: The values of the WSN data in the sliding window are changed by multiplying them by a constant -1. As a result, mirror data of the input data are generated, and the mirror imaging process for forming anomalies is also an injection method that focuses on helping the model learn the proportional data flow features.

3. Sudden change: Anomalies are created by greatly increasing or decreasing the WSN data in the sliding window, and the sudden change process is defined as shown (25), where $\tau$ is the duration of the anomaly, $X_{ij}(t)$ is the observation of the $i^{th}$ mode determined on node $j$ at time $t$, $X_i^{up}$ is the maximum observed value of the $i^{th}$ mode of the training set, and $X_i^{down}$ is the minimum value of the $i^{th}$ mode of the training set. To eliminate the impact of a small number of anomalies in the dataset, in this paper, $X_i^{up}$ is set to a larger value at the 99% quantile of the $i^{th}$ modal observation, and $X_i^{down}$ is set to a smaller value at the 1% quantile of the $i^{th}$ modal measurement. Through training, the model becomes able to distinguish between normal data and anomalies that spike or decrease suddenly, thereby increasing its sensitivity to data mutations.

$$X_{ij}(t) = X_{ij}(t) \pm \left(X_i^{up} - X_i^{down}\right) \quad t \in 1,2,...,\tau \quad (25)$$

4. Intermodal anomaly: An intermodal anomaly aims to find the parts of a positive or negative correlation that is present in multiple time series data modes and breaks the correlation between these modes. First, the user needs to set the start time $t_s$ and end time $t_e$ of the injection anomaly, and then, GLSL randomly generates a random node number *injnode* and a random mode number *injmodal*, and the anomaly is injected into the [$t_s$, $t_e$] fragment of the random mode of the random node. Then, GLSL separately calculates the correlation coefficients between the data flow of *injmodal* on *injnode* and the data flows of other modes on *injnode*. If the data flows are highly positively or negatively correlated, then we detect the trend of the data flow of *injmodal* on *injnode* and inject the anomaly that is opposite to this trend; specifically, we inject an anomaly with a gradually decreasing value when the data flow is in an uptrend, and vice versa. If none of the other modes are highly positively or negatively correlated with the *injmodal*, the intermodal anomaly injection process is abandoned, and a scale, negation or sudden change anomaly is randomly selected to be injected into [$t_s$, $t_e$] instead. The intermodal anomaly is detailed in Table I, where $p$ is the variable that controls the degree of deviation from normal data, and $\sigma$ is a minimal term that is much smaller than ($X_i^{up}$- $X_i^{down}$)/$p$. Assuming that $P$ and $Q$ are two known $n$-dimensional data vectors, the correlation coefficients of these two data vectors can be computed by (14), and each correlation coefficient is between −1 and 1.

$$\rho_{PQ} = Cov(P,Q) / \sqrt{Var[P]Var[Q]}$$

$$= \sqrt{\sum_{i=1}^{n}(P_i - \bar{P})^2 \cdot \sum_{i=1}^{n}(Q_i - \bar{Q})^2} \quad (25)$$

**Algorithm 1** Intermodal anomaly

**Input**: WSN data $X$, anomaly start time $t_s$, anomaly end time $t_e$.
**output**: WSN data after anomaly injection.
1:   Random initialize a injection node *injnode* and a injection modal *injmodal*
2:   **for** $modalnum = modal_0, modal_1, ..., modal_M$ **do**
3:     **if** $modalnum \neq injmodal$ **then**
4:       Obtain the data flow $X_{injmodal,injnode}[t_s:t_e]$ of the *injmodal* in the $t_s$-$t_e$ timestamps observed on the *injnode*
5:       Obtain the data flow $X_{modalnum,injnode}[t_s:t_e]$ of the *modalnum* in the $t_s$-$t_e$ timestamps observed on the *injnode*
6:       Calculate correlation coefficient $r$ between $X_{injmodal,injnode}[t_s:t_e]$ and $X_{modalnum,injnode}[t_s:t_e]$
7:       **if** $-1 < r < -0.8$ or $0.8 < r < 1$ **then**
8:         Check the trend of $X_{injmodal,injnode}[t_s:t_e]$
9:         **if** $X_{injmodal,injnode}[t_s:t_e]$ is on an upward trend **then**
10:           **for** $t$ from $t_s$ to $t_s+(t_e-t_s)/2$ **do**
11:             $X_{injmodal,injnode}(t) = X_{injmodal,injnode}(t-1)-(X_i^{up}-X_i^{down})/p\pm\sigma$
             # $\sigma$ is a small excitation, $\sigma \ll (X_i^{up}-X_i^{down})/p$, and $\sigma > 0$
12:           **for** $t$ from $t_s+(t_e-t_s)/2$ to $t_e$ **do**
13:             $X_{injmodal,injnode}(t) = X_{injmodal,injnode}(t-1)+(X_i^{up}-X_i^{down})/p\pm\sigma$
14:         **else**
15:           **for** $t$ from $t_s$ to $t_s+(t_e-t_s)/2$ **do**
16:             $X_{injmodal,injnode}(t) = X_{injmodal,injnode}(t-1)+(X_i^{up}-X_i^{down})/p\pm\sigma$
17:           **for** $t$ from $t_s+(t_e-t_s)/2$ to $t_e$ **do**
18:             $X_{injmodal,injnode}(t) = X_{injmodal,injnode}(t-1)-(X_i^{up}-X_i^{down})/p\pm\sigma$
19:         break
20:       **else**
21:         continue

**Algorithm 2** Internode anomaly

**Input**: WSN data $X$, anomaly start time $t_s$, anomaly end time $t_e$, the considered node number $k$, edge weight matrix $H$.
**output**: WSN data after anomaly injection.
1:   Random initialize a injection node *injnode* and a injection modal *injmodal*
2:   Find top $k$ neighbor nodes closest to *injnode* from the edge weight matrix to form the neighbor node set *Neibor*
3:   **for** $nodenum \in Neibor$ **do**
4:     **if** $nodenum \neq injnode$ **then**
5:       Obtain the data flow $X_{injmodal,injnode}[t_s:t_e]$ of the *injmodal* in the $t_s$-$t_e$ timestamps observed on the *injnode*
6:       Obtain the data flow $X_{injmodal,nodenum}[t_s:t_e]$ of the *injmodal* in the $t_s$-$t_e$ timestamps observed on the *nodenum*
7:       Calculate correlation coefficient $r$ between $X_{injmodal,injnode}[t_s:t_e]$ and $X_{injmodal,nodenum}[t_s:t_e]$
8:       **if** $-1 < r < -0.8$ or $0.8 < r < 1$ **then**
9:         Check the trend of $X_{injmodal,injnode}[t_s:t_e]$
10:         **if** $X_{injmodal,injnode}[t_s:t_e]$ is on an upward trend **then**
11:           **for** $t$ from $t_s$ to $t_s+(t_e-t_s)/2$ **do**
12:             $X_{injmodal,injnode}(t) = X_{injmodal,injnode}(t-1)-(X_i^{up}-X_i^{down})/p\pm\sigma$
13:           **for** $t$ from $t_s+(t_e-t_s)/2$ to $t_e$ **do**
14:             $X_{injmodal,injnode}(t) = X_{injmodal,injnode}(t-1)+(X_i^{up}-X_i^{down})/p\pm\sigma$
15:         **else**
16:           **for** $t$ from $t_s$ to $t_s+(t_e-t_s)/2$ **do**
17:             $X_{injmodal,injnode}(t) = X_{injmodal,injnode}(t-1)+(X_i^{up}-X_i^{down})/p\pm\sigma$
18:           **for** $t$ from $t_s+(t_e-t_s)/2$ to $t_e$ **do**
19:             $X_{injmodal,injnode}(t) = X_{injmodal,injnode}(t-1)-(X_i^{up}-X_i^{down})/p\pm\sigma$
20:         break
21:       **else**
22:         continue

5. Internode anomaly: An internode anomaly aims to find the parts of a positive or negative correlation that are present in multiple node time series data considering a single mode and breaks the correlation between such nodes. First, the user needs to set the start time $t_s$ and end time $t_e$ of the injection anomaly, as well as the considered number of nodes $k$. GLSL randomly generates a random node number *injnode* and a random mode number *injmodal*, and an anomaly is injected into the $[t_s, t_e]$ fragment of the random mode of the random node. Then, GLSL finds the $k$ neighboring nodes that are closest to *injnode* through the edge weight matrix and separately calculates the correlation coefficients between the data flow of *injmodal* on *injnode* and the data flows of *injmodal* on other neighboring nodes. If the data flows are highly positively or negatively correlated, then we detect the trend of the data flow of *injmodal* on *injnode* and inject the anomaly with the opposite trend; specifically, we inject an anomaly with a gradually decreasing value when the data flow is in an uptrend, and vice versa. If none of the other neighboring nodes are highly positively or negatively correlated with *injmodal*, the intermodal anomaly injection process is abandoned, and a scale, negation or sudden change anomaly is randomly selected to be injected into $[t_s, t_e]$ instead. The intermodal anomaly is detailed in Table I, where $p$ is the variable that controls the degree of deviation from normal data, and $\sigma$ is a minimal term that is much smaller than $(X_i^{up}-X_i^{down})/p$.

*B. Evaluation metrics*

We use the precision (Prec), recall (Rec) and F1 score (F1) metrics calculated over the test set to evaluate the performance of our method.

*Prec*: Precision can be used to measure the proportion of samples correctly predicted as abnormal by the proposed method out of the total number of samples predicted as abnormal.

*Rec*: The recall rate can be used to measure the proportion of samples correctly predicted as abnormal by the proposed method out of the total number of real abnormal samples.

*F1 score*: The F1 score considers both accuracy and recall. It is a metric defined according to the balance between these values.

*TP*, *TN*, *FP* and *FN* represent the four values of the confusion matrix. *TP* is the number of correctly detected anomalies (the predicted value $Y_{pred} = 1$, and the true label $L_t = 1$), *FP* is the number of incorrectly detected anomalies (the predicted value $Y_{pred} = 1$, and the true label $L_t = 0$), *TN* is the number of correctly identified normal values (the predicted value $Y_{pred} = 0$, and the true label $L_t = 0$), and *FN* is the number of undetected abnormal values (the predicted value $Y_{pred} = 0$, and the true label $L_t = 1$). $F1 = 2\times Prec \times Rec/(Prec+Rec)$, $Prec = TP/(TP+FP)$ and $Rec = TP/(TP+FN)$.

```
Algorithm 3    GLSL anomaly detection framework
Input: WSN data X, sliding window length W, timestamps number of train set
trainlength, timestamps number of train set testlength, injected anomaly duration τ.
Output: metrics of GLSL.
### train process
1:   for epochcount from 0 to epochnum do
2:       Initialize the sliding window start point winstart
3:       for winstart from 0 to trainlength-W do
4:           Obtain the WSN data within window X_now=X[:, :, winstart: winstart+W]
5:           Z=f_e(X_now;θ_e), X'=f_d(Z;θ_d), y=Softmax(FC(Z,X';θ_m))
6:           Calculate the loss value according to (21) and update [θ_e,θ_d,θ_m]
         according to the loss value.
7:   end for
### test process
8:   In the test set, T timestamps are sampled and 50% of them are divided into
     the injection anomaly detection point set S1, and the remaining 50% are
     divided into normal state detection point S2
9:   initialize TP=0, FN=0
10:  for checkpoint in S1 do
11:      Random initialize a injection node injnode and a injection modal injmodal
12:      According to Section V-B, inject one of the four types anomalies into the
         [checkpoint, checkpoint+τ] moment of the injmodal modal observed on the
         injnode node
13:      for winstart from 0 to testlength-W do
14:          Obtain the WSN data X_now=X[:, :, winstart: winstart+W]
15:          Z=f_e(X_now;θ_e), X'=f_d(Z;θ_d), y=Softmax(FC(Z,X';θ_m))
16:          if winstart+W in [checkpoint, checkpoint + delaystep] and y==1
17:              TP = TP + 1
18:              break
19:          if winstart+W > checkpoint+delaystep
20:              FN = FN + 1
21:              break
22:      end for
23:  end for
24:  initialize TN=0, FP=0
25:  for checkpoint in S2 do
26:      for winstart from 0 to testlength-W do
27:          Obtain the WSN data X_now=X[:, :, winstart: winstart+W]
28:          Z=f_e(X_now;θ_e), X'=f_d(Z;θ_d), y=Softmax(FC(Z,X';θ_m))
29:          if winstart+W == checkpoint
30:              if y==0
31:                  TN = TN + 1
32:                  break
33:              else
34:                  FP = FP + 1
35:                  break
36:      end for
37:  end for
38:  Calculate the metrics by TP、FN、TN、FP
```

Assuming that a sampling moment set with a length of $T$ is extracted from the test set, 50% of the sampling moments are extracted as set-1, and the remaining 50% are extracted as set-2. The moments in set-1 are the timestamps that need injected anomalies for testing, and the moments in set-2 are the timestamps that do not need injected anomalies for testing.

$T$ anomaly detection iterations are conducted on the entire test set. First, we detect the injection points of set-1. We randomly select one of the four methods mentioned in Section V-B every time to inject the corresponding anomaly into the test set data, and the injection position is a sampling timestamp of set-1. Assume that the sampling timestamp of an injection anomaly is $t$ and that the allowable delay constant is *delaystep*. If GLSL predicts the anomaly in [$t$, $t+delaystep$], GLSL correctly detects the injection anomaly (TP). If GLSL does not predict the anomaly in [$t$, $t+delaystep$], then there is an anomaly that is not detected (FN).

The allowable delay *delaystep* set in this paper is $W$; that is, the detection of an anomaly within a delay possessing the same value as the sliding window length is regarded as a successful detection. Then, the normal points in set-2 are detected; if GLSL does not predict an anomaly at the normal point with no injection anomaly, it is a correctly identified normal value (TN), and if an anomaly is predicted at a normal point, it is an incorrectly detected anomaly (FP). The overall GLSL approach is shown in algorithm 3.

*C. Sensitivity test*

We first train the model with a set of basic parameters and then test the sensitivity of the model to abnormal data (the impact of injecting abnormal data with different degrees of deviation from the normal data on the detection accuracy).

The experimental results of the sensitivity test are shown in Fig. 12, where $p$ is the deviation factor of the internode and intermodal anomalies mentioned in Section V-F, and when the value of $p$ increases, the degree of deviation of the anomaly injected at each timestamp from the normal data decreases. From the figure, it can be seen that as the deviation degree of the injection anomaly $p$ increases, the recall of the original model continues to decrease; in other words, when injecting more significant anomalies (when the $p$ value is small), the model more easily distinguishes between normal samples and abnormal samples, which means that the original model successfully captures the difference features between the original normal data and the abnormal data.

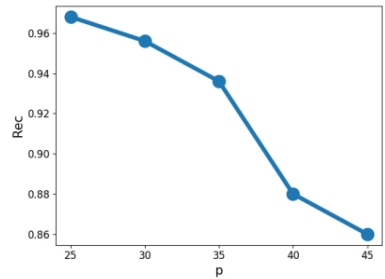

**Fig. 12.** Sensitivity test. When the value of $p$ increases, the degree of deviation of the anomaly injected at each timestamp from the normal data decreases.

After determining that the model designed in this paper has the ability to distinguish between normal data and abnormal data, in the following experiment, we modify the model parameters to fine-tune the model so that it can achieve better performance. The length of the sliding window $W \in$ [10,15,20,25,30], the learning rate $lr \in$ [0.00001,0.00005,0.0001,0.0005], the setting range of the GRU hidden layer is set as [8,16,32,64], the number of iteration epochs is 100, the deviation control variable $p$ for injecting anomalies is set to 40, and the duration ($t_e$-$t_s$) of the anomaly injection phase is set to 10. The framework in this article and the baselines in the next section are tested multiple times with the above parameter setting range, and the best results obtained across multiple tests are considered the final results and are displayed in the table.

## D. Baselines

In this section, this paper compares GLSL with three anomaly detection methods based on deep learning.

*1. CNN-LSTM:* This paper uses the PyTorch deep learning framework to build an 8-layer CNN-LSTM classification model with Conv2d → MaxPool2d → Conv2d → MaxPool2d → LSTM → FC → FC → FC. In each epoch, the framework aims to classify the normal data at times with no anomaly injection as class 0 and the data at times with anomaly injection as class 1. It is worth noting that the form of the 4-dimensional tensor input by Conv2d exactly matches the [batch, modul_num, node_num, window_size] form of the WSN data in this paper. However, the characteristics of the CNN make it unable to use the topology information of WSNs as a GNN does.

*2. MTAD:* This is an anomaly detection method that combines the advantages of prediction-based models and reconstruction-based models [7]. Prediction-based models are often used in application scenarios where historical data are used to predict the data at the next timestamp and can learn the period information of data, while reconstruction-based models can capture the overall distribution of entire time series datasets. MTAD uses two GATs to capture the correlation features between different modes and the temporal features between different timestamps, and then two types of features are spliced and fed into a GRU to solve the problem of long-term dependence. Finally, the network is divided into two branches to predict the data at the next timestamp and reconstruct the original input data. The inference score is composed of the prediction error and reconstruction error. When the inference score is higher than the preset threshold, the status at the current time is judged as abnormal. When abnormal data deviate from the normal data distribution, the inference score surges and exceeds the threshold.

It is worth noting that this method is only an anomaly detection method for multimodal data flows considering a single node, which is different from the anomaly detection method described in this paper for multinode multimodal data flows. This method does not consider the multinode situation and the spatial position features between sensor nodes; therefore, this paper trains 51 models on MTAD for the 51 sensor nodes in the dataset, and the performance metrics of MTAD shown in the table below are the average performance metrics obtained across these 51 models.

*3. Multinode multimodal time series anomaly detection method:* This is an anomaly detection method based on a prediction model [32], which is inspired by the previous baseline, MTAD. It solves the problem that MTAD does not consider multinode features and incorporates the spatial location information between multiple nodes into anomaly detection. The multinode multimodal time series anomaly detection method extracts the modal correlation features and historical time correlation features of different nodes through multiple GAT branches, solves the long-term dependence problem through a GRU, and finally predicts the detection values of all modes of all nodes at the next timestamp by capturing the spatial position features of the sensor nodes through a GAT. The inference score is calculated from the prediction error. When the inference score exceeds the preset threshold, the system status at the current time is judged as abnormal. When abnormal data deviate from the normal data distribution, the inference score surges and exceeds the threshold.

TABLE I
*Baselines*

| Method | Prec | Rec | F1 | Acc |
| --- | --- | --- | --- | --- |
| CNN-LSTM | 79.5% | 70.0% | 74.5% | 76.0% |
| MTAD | 77.5% | 86.0% | 81.5% | 80.5% |
| MMSAD | 89.1% | 82.0% | 85.4% | 86.0% |
| GLSL (GCN) | **97.6%** | 80.0% | 87.9% | 89% |
| GLSL (GAT) | 94.5% | **87.0%** | **90.6%** | **91.0%** |

The experimental results are shown in Table I. It can be found that the performance metric values of CNN-LSTM are lower than those of the other methods because a GNN can enable each node to extract interesting information from all neighboring nodes according to the WSN topology, which is more suitable than a convolutional layer that can only extract the interesting information within the range of the convolutional kernel. MTAD is a method that considers the correlations between multimodal data, but it does not consider the spatial location features of multiple sensor nodes for anomaly detection, so when using MTAD in a WSN with $N$ sensor nodes, it is necessary to train $N$ models for anomaly detection, and MTAD often triggers gradient explosion during the training process for some nodes, which is also a reason why MMSAD and GLSL are better than MTAD.

MMSAD is an anomaly detection method based on a prediction model. Although MMSAD uses a GNN and GRU to jointly capture the data features of different timestamps, when the ratio of the training set size to the test set size increases, the long-term dependency problem is still serious. When the distance between the prediction point and the dependent correlation information is far, it is difficult to learn the correlation information, resulting in low anomaly detection accuracy. GLSL combines the advantages of reconstruction-based models that can efficiently capture various features in normal data and self-supervised learning, which can improve the generalization ability of graph representation learning. The designed framework not only uses GNNs to extract more common information about the current nodes from a global graph perspective but also extracts special node information from the perspective of a single node. At the same time, it can apply existing GNNs and their variants to anomaly detection. The GCN kernel-based GLSL and GAT kernel-based GLSL approaches achieve the maximum accuracy and recall values, respectively. The comprehensive performance of the GAT kernel-based GLSL method is the best among all methods, with F1 score increases of 5.2% over MTAD and 10.5% over MMSAD.

## E. Interpretability of GLSL

To prove that the GLSL approach proposed in this paper can truly learn the distribution of normal data and capture the degree to which abnormal data deviate from normal data, we visualize the training and test data, the hidden layer $Z$ and the $X'$ of a sensor.

*Point anomaly test:* The raw data of 5000 moments of the 2nd mode observed on node 47 are shown in Fig. 13 (a), where the blue curve represents the data in the training set, and the orange curve represents the data in the test set. The green curve represents the prediction value, and the status at the current moment is predicted to be anomalous when the prediction value rises. It is worth noting that the prediction value in Fig. 13 (a) reaches 1 many times, which means that there are many misjudgments. However, this paper only assumes that the original data are all normal data; in fact, these raw data are not all normal data, and these misjudgment points can be used for analyzing the original data anomalies. Fig. 13 (b) shows the visualization produced by GLSL in the face of injected point anomalies. At the $500^{th}$ moment of the test set, we add a point anomaly that seriously deviates from the normal data value range, and correspondingly, it can be found that the judgment value increases during the lag period, which means that GLSL correctly identifies the injected point anomaly. Fig. 13 (c) shows the curve change produced by the hidden layer vector of the 2nd mode observed on node 47 in the test set before and after the injected point anomaly at the $511^{th}$ moment. The blue line is the hidden layer vector curve yielded when there is no injection anomaly, the red line is the hidden layer vector curve produced after anomaly injection, and a certain difference can be observed between the two shapes. Fig. 13 (d) shows the reconstruction curve change yielded by the 2nd mode observed on node 47 in the test set before and after the injected point anomaly at moment 510. The green line is the raw input curve of normal data because the value of the window length $W$ is set to 20. The green curve is actually the normal data in the moment range of 490-510. The blue curve is the reconstruction of the green curve, and the red curve is the reconstruction produced after an anomaly is injected into the normal input data. After injecting the point anomaly that significantly deviates from the normal numerical range, the reconstruction curve undergoes a large numerical change, and a sharp sudden change in the prediction curve occurs.

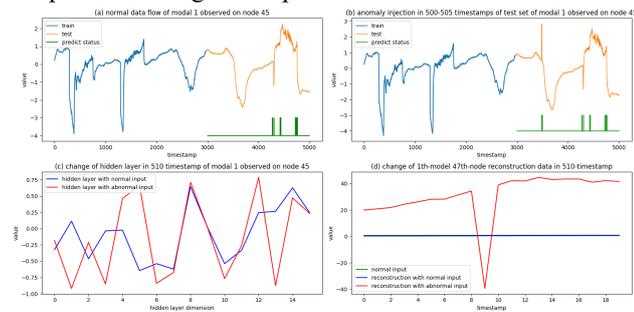

**Fig. 13.** Point anomaly test.

*Contextual anomaly test:* The raw data of 5000 moments of the 1st mode observed on node 45 are shown in Fig. 14 (a), where the blue curve represents the data in the training set, and the orange curve represents the data in the test set. The green curve represents the prediction values, and the status at the current moment is predicted to be anomalous when the prediction value rises. Fig. 14 (b) shows the visualization produced by GLSL in the face of injected contextual anomalies. The figure only shows the first modal data observed on node 45. In fact, a number of nodes adjacent to the $45^{th}$ node in this dataset are in declines similar to that of node 45 in moments 700-1000 of the test set. We add many values with small degrees of deviation from the normal data distribution at the $850^{th}$ moment of the test set, and the values do not exceed the interval of the normal data. Correspondingly, it can be found that the prediction value increases during the lag period, which means that GLSL identifies the injected context exception. Fig. 14 (c) shows the curve change exhibited by the hidden layer vector of the 1st mode observed on node 45 in the test set before and after the injected point anomaly at the $860^{st}$ moment. The blue line is the hidden layer vector curve yielded when there is no injection anomaly, the red line is the hidden layer vector curve produced after anomaly injection, and a certain difference can be observed between the two shapes. Fig. 14 (d) shows the reconstruction curve change produced by the 1st mode observed on node 45 in the test set before and after the injected point anomaly at moment 860. The green line is the raw input curve of the normal data because the value of the window length $W$ is set to 20. The green curve is actually the normal data in the moment range of 840-860. The blue curve is the reconstruction of the green curve, and the red curve is the reconstruction produced after an anomaly is injected into the normal input data. After injecting the contextual anomaly, the reconstruction curve changes numerically, and at the same time, a sharp sudden change occurs in the curve, but the magnitude of the numerical change is much smaller than that in Fig. 13 (d) because the context anomaly is still within the numerical interval of the normal data, and thus the significance of this anomaly is lower than that of the point anomaly.

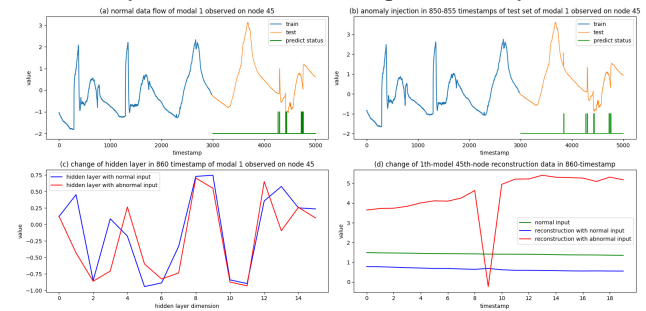

**Fig. 14.** Contextual anomaly test.

*Periodic anomaly test:* To test whether GLSL learns the periodic information contained in normal data during the training process, we inject a shape similar to those that appear in the training set into moments 150-180 and 1450-1480 of the 1st mode of the $45^{th}$ node in the test set to observe the judgment results of the model.

Fig. 15 shows the result of injecting a special shape into moments 150-180 of the test set. From Fig. 15 (b), we can find that because the interval of the special injected shape is at the junction of the two waveforms, this injection method obviously introduces an anomaly with a large deviation, and at the same time, GLSL also determines the abnormal state and identifies the injected anomaly within the corresponding lag period. Fig. 15 (c) shows the curve change yielded by the hidden layer vector of the 1st mode observed on node 45 in the test set before and after the injected point anomaly at time 183. Fig. 15 (d) shows the reconstruction curve change produced by the $1^{st}$ mode of node 45 in the test set before and after the injected point anomaly at time 183. Green represents the original normal input curve, blue signifies the reconstruction curve of the green curve without abnormality injection, and red denotes the reconstructed curve produced after injecting a special shape. We find that the reconstruction curve is similar to the original input curve when no injection abnormality occurs, but the reconstruction curve changes greatly after injecting a special shape, which means that GLSL successfully catches this injection anomaly.

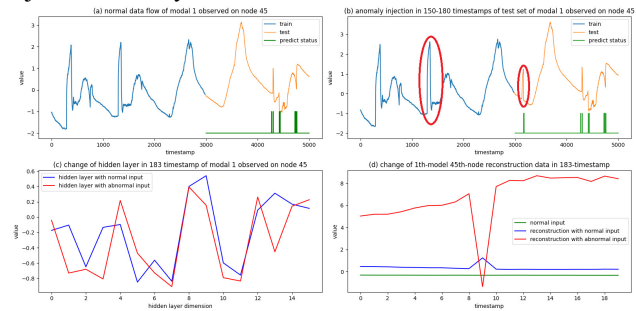

**Fig. 15.** Periodic anomaly test. The anomaly is injected into the test set at timestamps 150-180.

Fig. 16 shows the result of injecting a special shape into moments 1450-1480 of the test set. From Fig. 16 (b), we can find that since the special shape of the injection is similar to the previous curve of the training set for moments 400-600 and the injection position also conforms to the periodicity of the previous surge waveform, the model does not judge the special shape injected here as an anomaly. Combined with Fig. 15, adding this special waveform model at the junction of the waveform results in an abnormal judgment, and adding this special waveform similar to the previous waveform at this position in the test set results in a normal determination, which also shows that the model of the proposed framework successfully captures the period information of the WSN data flow. Fig. 16 (c) shows the curve change exhibited by the hidden layer vector of node 45 in the test set before and after the injected point anomaly at the 1483rd timestamp. Fig. 16 (d) shows the reconstruction curve change yielded by the 1st mode observed on node 45 in the test set before and after the injected point anomaly at timestamp 1483. The green curve is the original normal input curve, the blue curve is the reconstruction curve of the green curve without abnormality injection, and the red curve is the reconstruction curve produced after injecting a special shape. Although both reconstruction curves have certain deviations from the original input, this deviation is within a certain allowable range, so GLSL does not judge this moment as an anomaly.

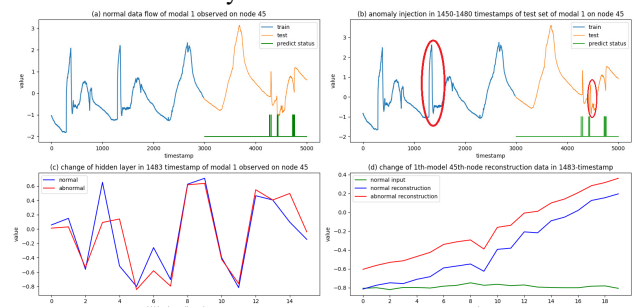

**Fig. 16.** Periodic anomaly test. The anomaly is injected into the test set at timestamps 1450-1480.

### F. GLSL$^+$ in large-scale WSN scenarios

When the network scale increase, the number of nodes and edges are increase exponentially, and the cost of training graph neural network model will also increase, this increases the computing cost and memory space exponentially. As mentioned in Section V-F, in order to cope with the problem of time consumption and hardware requirements caused by the large-scale of WSN network, an extended GLSL method named as GLSL+ is proposed based on K-means and PAA.

the performance of time complexsity on GLSL$^+$ by K-means and PAA are carried out through two sets of experiments. In experiment 1 senario, cluster the wireless sensor network while keep the dataset length constant, and different cluster runs GLSL anomaly detection framework on different devices, this reduces the parameter number of the constructed model, rationally reduce the GPU hardware overhead required for training and inference and the runtime overhead. In experiment 2 senario, PAA is used to reduce the dimensionality of the dataset in the time dimension while keep the clusters number constant, anomaly detection is performed again in a certain time interval (i.e. multiple moments) to reduce the number of runs to reduce time overhead.

TABLE II
LARGE-SCALE IMPROVING STRATEGY GLSL$^+$

| EXP | Cluster num | PAA | F1 | Avg running time |
|---|---|---|---|---|
| EXP. 1 | 1 | None | 90.6% | 19.0 |
|  | 2 | None | 89.9% | 17.8 |
|  | 3 | None | 87.9% | 16.0 |
|  | 4 | None | 85.7% | 14.3 |
| EXP. 2 | 3 | None | 87.9% | 16.0 |
|  | 3 | 4/5 | 89.3% | 15.4 |
|  | 3 | 3/5 | 80.2% | 11.7 |
|  | 3 | 2/5 | 71.7% | 7.8 |

As shown in Table II, we can see that with the growth of the class number, the comprehensive performance of the GLSL gradually decreases, but the average running time of the test set is also steadily decreasing, and when the size of the dataset is reduced from the original dimension to 4/5 of the original, the model shows better performance than before the dimensionality reduction, because with the dimensionality reduction of PAA, the general shape of the data has not changed, but some irregular points in the details are eliminated in the dimensionality reduction, and the training set is optimized so that the model is better trained and better performance is achieved.

However, as the degree of dimensionality reduction increased, the performance of the model began to deteriorate sharply, and when the degree of dimensionality reduction reached 2/5 of the original size, although the average running time dropped to more than twice the original size, the F1 value also dropped to 71.7%. According to the actual application requirements to weigh the importance between performance and average running time, selecting the appropriate number of classes number and PAA dimensionality reduction ratio can well apply GLSL to anomaly detection in large-scale wireless sensor network scenarios.

## VI. CONCLUSION

In this paper, an anomaly detection framework that integrates the advantages of unsupervised reconstruction model and self-supervised classification model is proposed, which can use any kind of graph neural network to adaptively fuse the common and special features of nodes, and make full use of the temporal features and spatial location features of WSN data to make state judgments. Experimental results show that the method can effectively detect abnormalities in the WSN in real time when injecting multiple types of abnormalities. However, the anomaly detection method proposed in this paper still needs to be processed on multiple modal data, although the modal number of WSN is often much smaller than the number of nodes, which means that the model size of the proposed method is less susceptible to node dimension expansion than MMSAD, but when the number of measured modals on each node continues to increase, the model constructed by this method will still continue to expand, so there is a defect of calculation speed for scenarios with a large number of modalities. In the next step, the algorithm is optimized by combining traditional anomaly detection methods, reducing hardware overhead, and reducing time complexity.